# Neuro Fuzzy Systems: State-of-the-art Modeling Techniques

Ajith Abraham

School of Computing & Information Technology
Monash University, Churchill 3842, Australia
http://ajith.softcomputing.net
Email: ajith.abraham@infotech.monash.edu.au

**Abstract:** Fusion of Artificial Neural Networks (ANN) and Fuzzy Inference Systems (FIS) have attracted the growing interest of researchers in various scientific and engineering areas due to the growing need of adaptive intelligent systems to solve the real world problems ANN learns from scratch by adjusting the interconnections between layers. FIS is a popular computing framework based on the concept of fuzzy set theory, fuzzy if-then rules, and fuzzy reasoning. The advantages of a combination of ANN and FIS are obvious. There are several approaches to integrate ANN and FIS and very often it depends on the application. We broadly classify the integration of ANN and FIS into three categories namely concurrent model, cooperative model and fully fused model. This paper starts with a discussion of the features of each model and generalize the advantages and deficiencies of each model. We further focus the review on the different types of fused neuro-fuzzy systems and citing the advantages and disadvantages of each model.

## 1. Introduction

Neuro Fuzzy (NF) computing is a popular framework for solving complex problems. If we have knowledge expressed in linguistic rules, we can build a FIS, and if we have data, or can learn from a simulation (training) then we can use ANNs. For building a FIS, we have to specify the fuzzy sets, fuzzy operators and the knowledge base. Similarly for constructing an ANN for an application the user needs to specify the architecture and learning algorithm. An analysis reveals that the drawbacks pertaining to these approaches seem complementary and therefore it is natural to consider building an integrated system combining the concepts. While the learning capability is an advantage from the viewpoint of FIS, the formation of linguistic rule base will be advantage from the viewpoint of ANN. In section 2 we present cooperative NF system and concurrent NF system followed by the different fused NF models in section 3. Some discussions and conclusions are provided towards the end.

## 2. Cooperative and Concurrent Neuro-Fuzzy Systems

In the simplest way, a cooperative model can be considered as a preprocessor wherein ANN learning mechanism determines the FIS membership functions or fuzzy rules from the training data. Once the FIS parameters are determined, ANN goes to the background. The rule based is usually determined by a clustering approach (self organizing maps) or fuzzy clustering algorithms. Membership functions are usually approximated by neural network from the training data.

In a concurrent model, ANN assists the FIS continuously to determine the required parameters especially if the input variables of the controller cannot be measured directly. In some cases the FIS outputs might not be directly applicable to the process. In that case ANN can act as a postprocessor of FIS outputs. Figures 1 and 2 depict the cooperative and concurrent NF models.

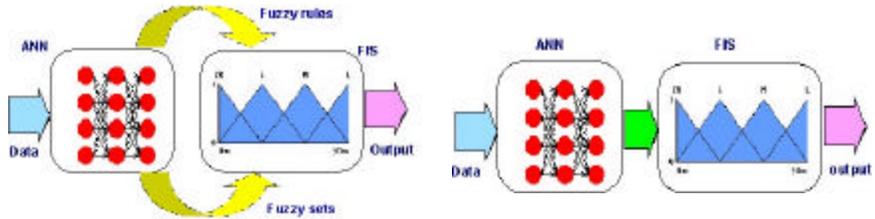 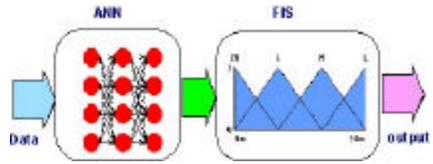

**Figure 1.** Cooperative NF model    **Figure 2.** Concurrent NF model

## 3. Fused Neuro Fuzzy Systems

In a fused NF architecture, ANN learning algorithms are used to determine the parameters of FIS. Fused NF systems share data structures and knowledge representations. A common way to apply a learning algorithm to a fuzzy system is to represent it in a special ANN like architecture. However the conventional ANN learning algorithms (gradient descent) cannot be applied directly to such a system as the functions used in the inference process are usually non differentiable. This problem can be tackled by using differentiable functions in the inference system or by not using the standard neural learning algorithm. Some of the major woks in this area are GARIC [9], FALCON [8], ANFIS [1], NEFCON [7], FUN [3], SONFIN [2], FINEST [4], EFuNN [5], dmEFuNN[5], evolutionary design of neuro fuzzy systems [10], and many others.

- **Fuzzy Adaptive learning Control Network (FALCON)**

FALCON [8] has a five-layered architecture as shown in Figure 3. There are two linguistic nodes for each output variable. One is for training data (desired output) and the other is for the actual output of FALCON. The first hidden layer is responsible for the fuzzification of each input variable. Each node can be a single node representing a simple membership function (MF) or composed of multilayer nodes that compute a complex MF. The Second hidden layer defines the preconditions of the rule followed by rule consequents in the third hidden layer. FALCON uses a hybrid-learning algorithm comprising of unsupervised learning to locate initial membership functions/ rule base and a gradient descent learning to optimally adjust the parameters of the MF to produce the desired outputs.

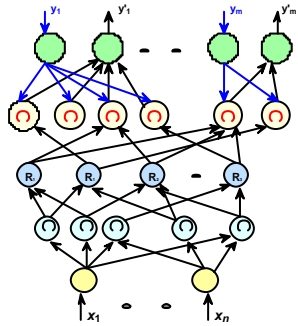 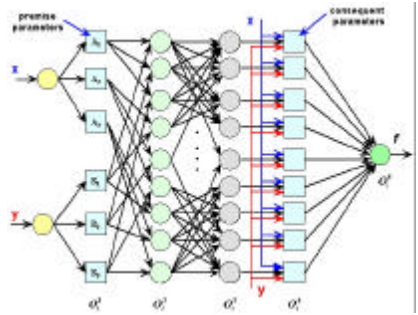

**Figure 3. Architecture of FALCON**     **Figure 4.** Structure of ANFIS

- **Adaptive Neuro Fuzzy Inference System (ANFIS)**

ANFIS [1] implements a Takagi Sugeno FIS and has a five layered architecture as shown in Figure 2. The first hidden layer is for fuzzification of the input variables and T-norm operators are deployed in the second hidden layer to compute the rule antecedent part. The third hidden layer normalizes the rule strengths followed by the fourth hidden layer where the consequent parameters of the rule are determined. Output layer computes the overall input as the summation of all incoming signals. ANFIS uses backpropagation learning to determine premise parameters (to learn the parameters related to membership functions) and least mean square estimation to determine the consequent parameters. A step in the learning procedure has got two parts: In the first part the input patterns are propagated, and the optimal consequent parameters are estimated by an iterative least mean square procedure, while the premise parameters are assumed to be fixed for the current cycle through the training set. In the second part the patterns are propagated again, and in this epoch, backpropagation is used to modify the premise parameters, while the consequent parameters remain fixed. This procedure is then iterated.

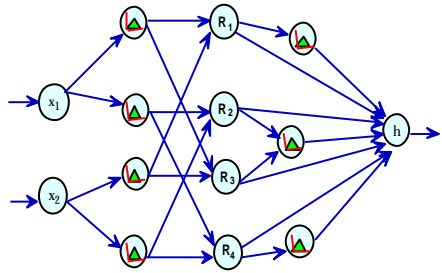 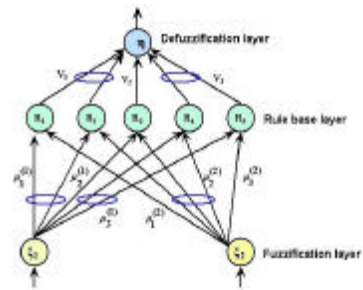

**Figure 5.** ASN of GARIC     **Figure 6.** Architecture of NEFCON

- **Generalized Approximate Reasoning based Intelligent Control (GARIC)**

GARIC [9] implements a neuro-fuzzy controller by using two neural network modules, the ASN (Action Selection Network) and the AEN (Action State Evaluation

Network). The AEN is an adaptive critic that evaluates the actions of the ASN. ASN of GARIC is feedforward network with five layers. Figure 5 illustrates the structure of GARIC – ASN. The connections between layers are not weighted. The first hidden layer stores the linguistic values of all the input variables. Each input unit is only connected to those units of the first hidden layer, which represent its associated linguistic values. The second hidden layer represents the fuzzy rules nodes, which determine the degree of fulfillment of a rule using a *softmin* operation. The third hidden layer represents the linguistic values of the control output variable *?*. Conclusions of the rule are computed depending on the strength of the rule antecedents computed by the rule node layer. GARIC makes use of local mean-of-maximum method for computing the rule outputs. This method needs a crisp output value from each rule. Therefore the conclusions must be defuzzified before they are accumulated to the final output value of the controller. GARIC uses a mixture of gradient descent and reinforcement learning to fine-tune the node parameters.

- **Neuro-Fuzzy Control (NEFCON)**

NEFCON [7] is designed to implement Mamdani type FIS and is illustrated in Figure 6. Connections in NEFCON are weighted with fuzzy sets and rules ($\mu$, *?* are the fuzzy sets describing the antecedents and consequents) with the same antecedent use so-called shared weights, which are represented by ellipses drawn around the connections. They ensure the integrity of the rule base. The input units assume the task of fuzzification interface, the inference logic is represented by the propagation functions, and the output unit is the defuzzification interface. The learning process of the NEFCON model is based on a mixture of reinforcement and backpropagation learning. NEFCON can be used to learn an initial rule base, if no prior knowledge about the system is available or even to optimize a manually defined rule base. NEFCON has two variants: NEFPROX (for function approximation) and NEFCLASS (for classification tasks) [7].

- **Fuzzy Inference and Neural Network in Fuzzy Inference Software (FINEST)**

FINEST [4] is capable of two kinds of tuning process, the tuning of fuzzy predicates, combination functions and the tuning of an implication function. The generalized modus ponens is improved in the following four ways (1) Aggregation operators that have synergy and cancellation nature (2) A parameterized implication function (3) A combination function that can reduce fuzziness (4) Backward chaining based on generalized modus ponens. FINEST make use of a backpropagation algorithm for the fine-tuning of the parameters. Figure 7 shows the layered architecture of FINEST and the calculation process of the fuzzy inference. FINEST provides a framework to tune any parameter, which appears in the nodes of the network representing the calculation process of the fuzzy data if the derivative function with respect to the parameters is given.

- **FUzzy Net (FUN)**

In FUN [3], the neurons in the first hidden layer contain the membership functions and this performs a fuzzification of the input values. In the second hidden layer, the conjunctions (fuzzy-*AND*) are calculated. Membership functions of the output variables are stored in the third hidden layer. Their activation function is a fuzzy-*OR*. Finally the output neuron performs the defuzzification. The network is initialized with a fuzzy rule base and the corresponding membership functions and there after uses a stochastic learning technique that randomly changes parameters of membership functions and connections within the network structure. The learning process is driven by a cost function, which is evaluated after the random modification. If the modification resulted in an improved performance the modification is kept, otherwise it is undone.

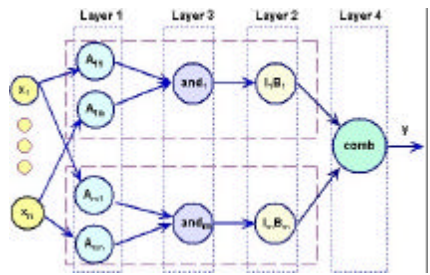
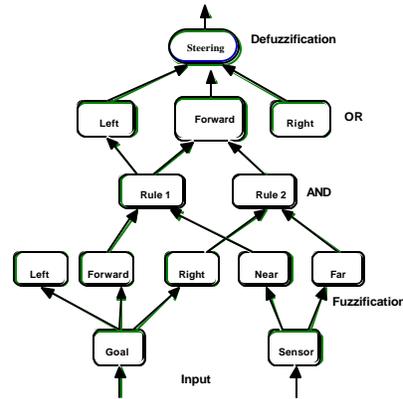

**Figure 7**. Architecture of FINEST   **Figure 8**. Architecture of FUN

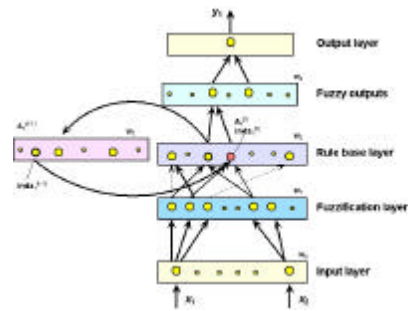
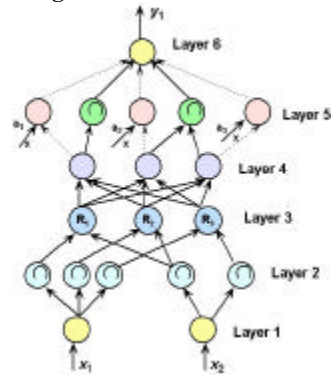

**Figure 9.** Architecture of EFuNN   **Figure 10.** Architecture of SONFIN

- **Evolving Fuzzy Neural Network (EFuNN)**

In EFuNN [5] all nodes are created during learning. The input layer passes the data to the second layer, which calculates the fuzzy membership degrees to which the input values belong to predefined fuzzy membership functions. The third layer contains fuzzy rule nodes representing prototypes of input-output data as an association of hyper-spheres from the fuzzy input and fuzzy output spaces. Each rule node is defined by 2 vectors of connection weights, which are adjusted through the hybrid learning technique. The fourth layer calculates the degrees to which output membership functions are matched by the input data, and the fifth layer does defuzzification and calculates exact values for the output variables. Dynamic Evolving Fuzzy Neural Network (dmEFuNN) [5] is a modified version of EFuNN with the idea that not just the winning rule node's activation is propagated but a group of rule nodes is dynamically selected for every new input vector and their activation values are used to calculate the dynamical parameters of the output function. While EFuNN implements fuzzy rules of Mamdani type, dmEFuNN estimates the Takagi-Sugeno fuzzy rules based on a least squares algorithm.

- **Self Constructing Neural Fuzzy Inference Network (SONFIN)**

SONFIN [2] implements a modified Takagi-Sugeno FIS and is illustrated in Figure 10. In the structure identification of the precondition part, the input space is partitioned in a flexible way according to an aligned clustering based algorithm. As to the structure identification of the consequent part, only a singleton value selected by a clustering method is assigned to each rule initially. Afterwards, some additional significant terms (input variables) selected via a projection-based correlation measure for each rule are added to the consequent part (forming a linear equation of input variables) incrementally as learning proceeds. For parameter identification, the consequent parameters are tuned optimally by either least mean squares or recursive least squares algorithms and the precondition parameters are tuned by backpropagation algorithm.

- **Evolutionary Design of Neuro-Fuzzy Systems**

In the evolutionary design of NF systems [10], the node functions, architecture and learning parameters are adapted according to a five-tier hierarchical evolutionary search procedure as shown in Figure 11(b). The evolving NF model can adapt to Mamdani or Takagi Sugeno type FIS. Only the layers are defined in the basic architecture as shown in Figure 11(a). The evolutionary search process will decide the optimal type and quantity of nodes and connections between layers. Fuzzification layer and the rule antecedent layer functions similarly to other NF models. The consequent part of rule will be determined according to the inference system depending on the problem type, which will be adapted accordingly by the evolutionary search mechanism. Defuzzification/ aggregation operators will also be adapted according to the FIS chosen by the evolutionary algorithm. Figure 11(b) illustrates the computational framework and interaction of various evolutionary search procedures. For every learning parameter, there is the global search of inference mechanisms that proceeds on a faster time scale in an environment decided by the inference system and the problem. For every inference mechanism there is the global search of fuzzy rules (architecture) that proceeds on a faster time scale in an environment decided by the learning parameters, inference system and the problem. Similarly, for every architecture, evolution of membership function parameters proceeds at a faster time scale in an environment decided by the architecture, inference mechanism, learning rule, type of inference system and the problem. Hierarchy of the different adaptation procedures will rely on the prior knowledge. For example, if there is more prior knowledge about the architecture than the inference mechanism then it is better to implement the architecture at a higher level.

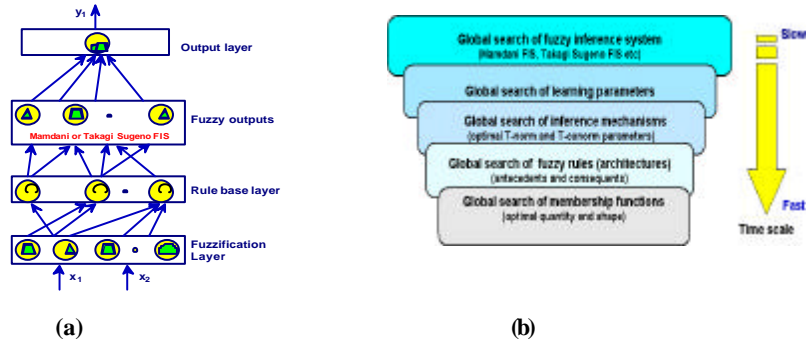

**(a)**          **(b)**

**Figure 11(a).** Architecture and **(b)** computational framework for evolutionary design of neuro fuzzy systems

## 4. Discussions

As evident, both cooperative and concurrent models are not fully interpretable due to the presence of ANN (black box concept). Whereas a fused NF model is interpretable and capable of learning in a supervised mode. In FALCON, GARIC, ANFIS, NEFCON, SONFIN, FINEST and FUN the learning process is only concerned with parameter level adaptation within fixed structures. For large-scale problems, it will be too complicated to determine the optimal premise-consequent structures, rule numbers etc. User has to provide the architecture details (type and quantity of MF's for input and output variables), type of fuzzy operators etc. FINEST provides a mechanism based on the improved generalized modus ponens for fine tuning of fuzzy predicates & combination functions and tuning of an implication function. An important feature of EFuNN and dmEFuNN is the one pass (epoch) training, which is highly capable for online learning. Since FUN system uses a stochastic learning procedure, it is questionable to call FUN a NFy system. As the problem become more complicated manual definition of NF architecture/parameters becomes complicated. Especially for tasks requiring an optimal NF system, evolutionary design approach might be the best solution. Table 1 provides a comparative performance [11] of some neuro fuzzy systems for predicting the Mackey-Glass chaotic time series [6]. Training was done using 500 data sets and NF models were tested with another 500 data sets.

**Table 1.** Performance of NF systems and ANN

| System | Epochs | RMSE |
|---|---|---|
| ANFIS | 75 | 0.0017 |
| NEFPROX | 216 | 0.0332 |
| EFuNN | 1 | 0.0140 |
| dmEFuNN | 1 | 0.0042 |
| SONFIN | - | 0.0180 |

## 5. Conclusions

In this paper we have presented the state of art modeling of different neuro-fuzzy systems. Due to the lack of a common framework it remains often difficult to compare the different neuro-fuzzy models conceptually and evaluate their performance comparatively. In terms of RMSE error, NF models using Takagi Sugeno FIS performs better than Mamdani FIS even though it is computational expensive. As a guideline, for NF systems to be highly intelligent some of the major requirements are fast learning (memory based - efficient storage and retrieval capacities), on-line adaptability (accommodating new features like inputs, outputs, nodes, connections etc), achieve a global error rate and computationally inexpensive. The data acquisition and preprocessing training data is also quite important for the success of neuro-fuzzy systems. All the NF models use gradient descent techniques to learn the membership function parameters. For faster learning and convergence, it will be interesting to explore other efficient neural network learning algorithms (e.g. conjugate gradient search) instead of backpropagation.